# Empirical Study of Overfitting in Deep FNN Prediction Models for Breast Cancer Metastasis

Chuhan Xu, Pablo Coen-Pirani, Xia Jiang

*Abstract*—Overfitting is defined as the fact that the current model fits a specific data set perfectly, resulting in weakened generalization, and ultimately may affect the accuracy in predicting future data. In this research we used an EHR dataset concerning breast cancer metastasis to study overfitting of deep feedforward Neural Networks (FNNs) prediction models. We included 11 hyperparameters of the deep FNNs models and took an empirical approach to study how each of these hyperparameters was affecting both the prediction performance and overfitting when given a large range of values. We also studied how some of the interesting pairs of hyperparameters were interacting to influence the model performance and overfitting. The 11 hyperparameters we studied include activate function; weight initializer, number of hidden layers, learning rate, momentum, decay, dropout rate, batch size, epochs, L1, and L2. Our results show that most of the single hyperparameters are either negatively or positively corrected with model prediction performance and overfitting. In particular, we found that overfitting overall tends to negatively correlate with learning rate, decay, batch sides, and L2, but tends to positively correlate with momentum, epochs, and L1. According to our results, learning rate, decay, and batch size may have a more significant impact on both overfitting and prediction performance than most of the other hyperparameters, including L1, L2, and dropout rate, which were designed for minimizing overfitting. We also find some interesting interacting pairs of hyperparameters such as learning rate and momentum, learning rate and decay, and batch size and epochs.

## I. Introduction

Our results are consistent with some of the existing research finding or knowledge such as activation function is associated with overfitting, the predict performance tends to drop when a lot of momentum works together with a large learning rate, and a smaller batch size is often associated with better prediction performance. Our results not only substantiate some of the existing knowledge in the field of machine learning but also present interesting new findings. These types of findings are useful for mitigating overfitting when conducting hyperparameter tuning and selecting the range of hyperparameter values required by grid search.

### A. Breast Cancer

Breast cancer was responsible for 685,000 deaths worldwide in 2022, and with no direct cure, it will remain one of the main cancer-associated causes of death in women globally for the foreseeable future [1]-[2]. Breast cancer is also the number one cause of cancer-related deaths for US women aged 20 to 59, estimated to account for 43600 deaths in 2021 [3]-[6]. Breast cancer metastasis is the main cause of breast cancer death [7]. Being able to effectively predict the likelihood of metastatic occurrence for each individual patient is important because the prediction can guide treatment plans tailored to a specific patient to prevent metastasis and to help avoid under-or over-treatment.

### B. Deep Learning

Deep Machine learning methods are now more and more used in health care related prediction tasks including the prediction of breast cancer metastasis [8]. Improving the prediction performance of these methods has been a key effort in the field of biomedical informatics and artificial intelligence in health care [9]. A Neural Network (NN) is one of the machine learning methods that can be used to conduct prediction. ANN consists of layers of artificial neurons, also called nodes, mimicking loosely how human brain and nervous system work to process and pass signals via neurons [10]. Due to this, an ANN is also referred to as an Artificial Neural Network (ANN). A traditional ANN consists of one input layer, one hidden layer, and one output layer [11]. Deep learning is the use of a non-conventional ANN that is composed of more than one hidden layer, which is also referred to as a Deep Neural Network (DNN) [12].

### C. Grid Search

In a grid search, each of the hyperparameters is given a preselected series of values, the program will then iterate through every hyperparameter value combination possible to train models. In addition to Grid search, there are other approaches of hyperparameter tunning including Bayesian optimization and Genetic Algorithm. One of the advantages of grid search is that the hyperparameter settings are independent. This makes it suitable to conduct parallel computing.

Manuscript received August 3, 2022, This work was supported by the U.S. Department of Defense through the Breast Cancer Research Program under Award No. W81XWH1910495 (to XJ). Other than supplying funds, the funding agencies played no role in the research.

The authors are with the Department of Biomedical Informatics, University of Pittsburgh, PA 15217 USA (e-mail: xij6@pitt.edu).

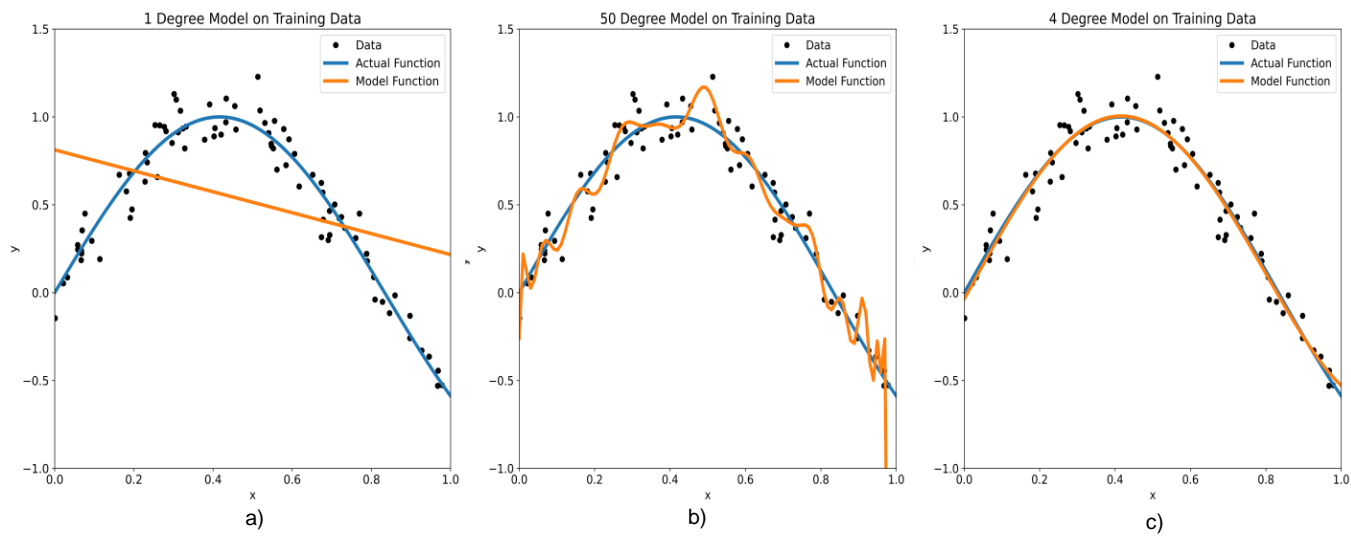

Fig. 1. Underfitting and overfitting on randomized data.

## D. Overfitting

Overfitting is a very common problem in deep learning and usually has a negative impact on the prediction performance of deep learning. A Google scholar search using the key word "overfitting in deep learning" returns more than 280000 relevant publications. The growth trend of the number of articles about overfitting in deep learning during the past 5 years, based on the Google Scholar data, illustrates that research interests in overfitting concerning deep learning increased dramatically from year 2017 to year 2020, with relevant research articles jumped from 18200 to 82700 within three years. This may also reflect the importance of handling overfitting in deep learning.

Overfitting occurs when the model performs well on training data but generalizes poorly to unseen data. The reasons for this include a limited size of the training dataset, imbalance in the training dataset, the complexity of models, and so on [13]. For example, in Fig. 1a) and b) there is a generated data set that has been used to train a polynomial model of varying degree. Fig. 1 b) is an example of the model overfitting because the polynomial function's degree is too high. Since the function is long and overly complex it fits to the data too well and begins directly connecting the data points. This is bad because when the model is run on data that it has never seen before, it will predict incorrectly based on the training data and not the testing data.

While overfitting is the subject of this paper and arguably more common, its antithesis is also important to consider when training and testing new models [14]. Underfitting is a problem where the model performs badly on both training data and unseen data, generally occurring when a model is overly regularized, inadequately trained, or lacks relevant predictive features [15]. Fig. 1a) is clearly underfitted as the model has not captured the trend in the data accurately. To fix this, the degree must be raised, but if it is too high then the model is too complex and overfitting occurs, as seen in Fig. 1b). By tuning the degree parameter, the model in Fig. 1c) below can be obtained, which is the optimized function in this case. When looking at an underperforming model, it is important to distinguish between the model being outright incorrect and it being underfitted. For example, if a model is trained for too long and is too complex, the next logical step is to remove features and reduce training time as said model is overfitted. As can be seen from Fig. 1a), if these logical steps go too far, it will turn into an underfitted model as a consequence of attempting to prevent overfitting.

## II. OVERFITTING AND HYPERPARAMETERS

Another example of overfitting and underfitting can be found within this paper: when looking at a graph containing epochs as a hyperparameter, such as Fig. 2. "Epochs" defines the number of times that the entire training data are used by the learning algorithm during training. One epoch means every sample in the training set has been used exactly once to update the internal model parameters. As seen in Fig. 2, the first few epochs will either be emitted or look vastly incorrect, in this case the latter. These sorts of results are a direct case of underfitting as the deep learning model needs multiple full passes of the data to find the pattern and adapt adequately. Additionally, the higher the epochs the further mean_train_AUC and mean_test_AUC get from each other, which means we are encountering overfitting near the end of this graph. This is because the model is performing better and better on training data but worse and worse on testing data. While in some cases the model was underfitted, in most cases where both the training and test data performance is bad it will turn out to be a problem with the model itself.

In practice, underfitting can be mostly avoided by increasing the level of training, but it is sometimes challenging to strike a balance between training sufficiently and training too much (overfitting). This is the reason this paper focuses on overfitting — it is much harder to manage overfitting within deep learning model which happens more frequently than underfitting as deep learning networks in the medical field get increasingly complex. In most cases, deep learning model has more hyperparameters than other normal machine learning models which also illustrates the complexity.

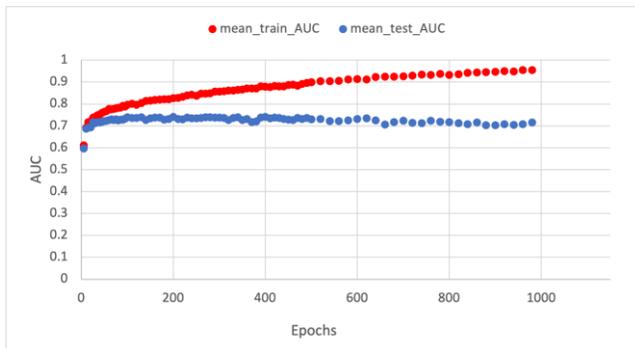

Fig. 2. Training epoch with AUCs.

It is not possible to completely eliminate overfitting in deep learning, but there are known hyperparameters of deep learning which can be adjusted to reduce the effect of overfitting [16]. As mentioned previously, "epochs", a hyperparameter that helps balance model convergence, is a known factor of overfitting in deep learning [17]. The "dropout rate" is another hyperparameter that is known to affect overfitting [18]. Neurons are randomly selected and dropped out during training based on the preselected dropout rate to reduce time cost and minimize model overfitting. Additionally, regularization hyperparameters L1 and L2 are known to reduce overfitting [19]. L1, also called a sparsity regularization factor, is a factor that can be used to remove the effect of the "noisy" input nodes and make the network less complex. L2 is a regularization factor based on weight-based decay, which penalizes large weights to adjust the weight updating step during model training. Activation function also plays an important role in preventing the model from overfitting [20]. The role of the activation function in neural networks is to introduce nonlinear factors with nonlinear combinations of weighted inputs to solve problems that cannot be solved by linear models including Sigmoid activation function, ReLu activation function and so on. Finally, number of hidden layers is another factor that may impact the overfitting. Every neural network will include multiple layers, with each layer of neurons receiving input from the previous layer and producing output to the next layer, so a large number of hidden layers may lead to overfitting problem [21].

Deep learning became a popular method due to its success in a lot of image-relevant applications, and part of its success is attributed to its various hyperparameters that can be optimized [22]. The process of optimizing the hyperparameters is also called hyperparameter tuning, which often involves selecting the set of hyperparameter values that has the best prediction performance out of all sets of hyperparameter values being tested [23]. The sets of hyperparameter values can either be manually selected or selected automatically following certain rules, and the latter method is often called grid search. Grid search is designed to conduct hyperparameter tuning in a systematic way by going through each of the sets of hyperparameter values automatically during the model training process [23]. Other than epochs, dropout rate, L1, and L1, the four hyperparameters that are known to affect overfitting, various other hyperparameters can be tuned in a grid search of deep learning [24]. The hyperparameter "optimizer" can assume different values such as SGD (Stochastic Gradient Descent) and Adagrad (Adaptive Gradient Descent). SGD adjusts its learning rate via "momentum" and "decay", the two other hyperparameters that can be tuned via grid search. The momentum, a moving average of the gradients that can help accelerate the convergence of training. The decay is an iteration-based decay factor that can be used to decrease learning rate in each epoch during the optimization process [25]. The "learning rate" is a hyperparameter that governs how big of a step it takes each time to update the internal model parameters (weights and biases) in response to the estimated error during the model training process [26]. It is used by both the SGD and Adagrad. Adagrad adapts the learning rate to the parameters, conducting smaller-step updates for parameters linked to frequently appearing features, and larger-step updates for parameters linked to less frequent features. The batch size is also a hyperparameter of deep learning, which controls the number of the training samples that are "fed" to the neural network before internal model parameters are updated [27].

In a grid search, each of the hyperparameters is given a preselected series of values, the program will then iterate through every hyperparameter value combination possible to train models [23]. We call a hyperparameter value combination a hyperparameter setting. There is great uncertainty when choosing a set of values for a hyperparameter when conducting a grid search. For example, the traditional textbook or default value for learning rate is 0.01. But technically speaking learning rate can assume numerous values. So, should we choose 0.001 to 0.01 with a step size of 0.001 or should we choose 0.0001 to 0.05 with a step size of 0.005? In addition, there is no definite answer as to whether and how much learning rate affects overfitting in deep learning.

Although some of the hyperparameters such as epochs, dropout rate, L1, and L2 are known to have an influence on overfitting, there are still questions such as to which of them has the least or the largest effect. Similarly to learning rate, there is still a lot of uncertainty when selecting a range of values for such a hyperparameter, which is required by grid search. For instance, since technically there is no upper bound for the value of number of epochs. You can choose 1 to 500 epochs or 1 to 5000 epochs when doing grid search. As previously illustrated, when the number of epochs is too low, you can underfit and when it is too high, you can overfit. Furthermore, when the number of epochs get higher, grid search becomes slower, and you can waste a lot of computing time with worse results due to overfitting. So, for these hyperparameters what are the ranges of values that you can choose to most likely avoid underfitting and overfitting?

Most of the research on deep learning in the medical field is based on image identification focusing on convolutional neural networks [28]. Empowered with large scale neural networks and massively parallel computing devices, the accuracy of image recognition is greatly improved [22]. However, image data are only one type of "Big Data". There exist a large quantity and variety of non-image data that can be very valuable to machine learning and personalized medicine [29]. For instance, the electronic health record (EHR), a widely available data resource, can be utilized for the purpose of tailoring therapies and providing prognostic information. An EHR database contains abundant data about patients' clinical features, disease status, interventions, and clinical outcomes. Such data are invaluable to tailoring diagnosis and prognoses to individual with diseases such as breast cancer [30]-[31]. In this research we used a EHR dataset concerning breast cancer metastasis to study overfitting of deep feedforward Neural Networks (FNNs) prediction models [31]. We included 11 hyperparameters of the deep FNNs models and took an empirical approach to study how each of these hyperparameters was affecting both the prediction performance and overfitting when given a large range of values. We also studied how some of the interesting pairs of hyperparameters were interacting to influence the model performance and overfitting. We hope, through our study, to help answer some of the questions as mentioned in the previous paragraphs. We hope to obtain interesting findings that add to our exiting knowledge about overfitting and be helpful to the grid search approach of learning, or at least stir research interests in this direction.

## III. EXPERIMENTS METHOD

In this research, we conducted a unique type of grid search with deep learning repeatedly to study how each of the 11 hyperparameter influences model overfitting and prediction performance when assuming various values. In this type of grid search, we gave a wide range of values to the hyperparameter being studied at the time while each of the other 10 hyperparameters assumed a single value, which was randomly picked from a set of values. For each of the 11 hyperparameters, we repeated this type of grid search 30 times, so that we can get the mean measurements of model overfitting and prediction performance, averaged over 30 values, for each of the range of values given to the hyperparameter of interest. Next, we discuss in detail all aspects of the method we used.

### A. Feedforward Deep Neural Networks(FNNs)

The deep neural networks used in this study are fully connected FNNs that have at least one hidden layer [23]. Inspired by biology, these neural networks do not contain cycles and each data point will simply traverse a chain of hidden layers. Fig. 3 contains a summary of the inner workings of one of the FNNs that we developed in this study. It includes four hidden layers and two output layers. The 31 input nodes to this neural network represent the 31 clinical features contained in the patient data that will help the model predict breast cancer metastasis, and the output layer contains two nodes representing the binary status of breast cancer metastasis. Each node in the model has an activation function, represented by f(x), which decides the node's individual output value established by the current value of the node. In both the input layer and the hidden layer(s), we employed a rectifier linear unit (ReLU) [37] as the activation function. In Fig. 3 each hidden layer has a certain number of hidden nodes that can be different from the other layers, as can be seen by the distinct variables p, q, m, and r, respectively. Initialize the weight matrix of neural network using glorot_normal initializer. This FNN model was developed in Python using the TensorFlow and Keras packages.

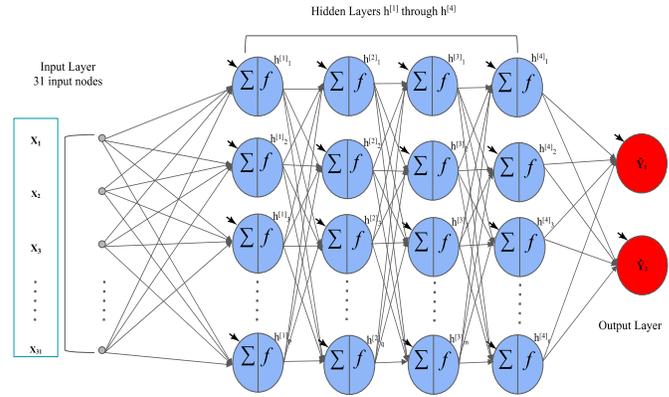

Fig. 3. The structure of a FNN model that we developed.

### B. Unique Types of Grid Search Designed for this Study

#### 1) Single hyperparameter

We considered 11 hyperparameters including activation function, weight initializers, learning rate, momentum, decay, dropout rate, epochs, batch size, L1 and L2 in this study. For each of the 11 hyperparameters we conducted a unique type of grid search with deep learn 30 times, and each time we gave a wide range of values to the one hyperparameter of interest while each of the remaining hyperparameters assumes its base value. The base values of the hyperparameters came from the best model resulted from grid searches in our previous study [23]. Table 1 shows the hyperparameters, their base values, and the values that we tested in this research for the purpose of studying overfitting and underfitting.

TABLE I
HYPERPARAMETERS AND THEIR VALUES IN GRID SEARCH

| Hyperparameters (number of values tested) | Values Tested | Base Values |
|---|---|---|
| 1.Activation Function for output layer (2) | Sigmoid<br>Hard Sigmoid | Sigmoid |
| 2.Weight Initializers(5) | Uniform<br>Normal<br>Glorot Uniform<br>Glorot Normal<br>Lecun Uniform | Glorot_Normal |
| 3.Number of hidden layers (4) | 1 layer (75,1)<br>2 layer (75, 75, 1)<br>3 layer (75, 75, 75, 1)<br>4 layer (75, 75, 75, 75, 1) | 2 |
| 4.Learning Rate(400) | 0.001- 0.4 with the step size 0.001 | 0.005 |
| 5.Momentum Beta(9) | 0.1-0.9 with step size 0.1 | 0.9 |
| 6.Iteration-based decay(400) | 0-0.2 with step size 0.0005 | 0.01 |
| 7.Dropout Rate(9) | 0.1-0.9 with step size 0.1 | 0.5 |
| 8.Training Epochs(80) | 5-2000 with step size 25 | 100 |
| 9.Batch size(838) | 1-4186(dataset size) with step size 5 | 10 |
| 10.L1(501) | 0-0.5 with step size 0.001 | 0 |
| 11.L2(701) | 0-0.7 with step size 0.001 | 0.008 |

### 2) Paired hyperparameters

We conducted a grid search in which we gave a range of values to a pair of hyperparameters while fixing the values of the remaining hyperparameters, which was sometimes referred to as an interactive grid search in this text. When a lot of momentum combines with high learning rate, training can change in large steps to pass by the global minimum. It is therefore believed that a setup where learning rate and momentum have a negative correlation is the best [40]. Due to this we are interested in knowing how learning rate and momentum interact to affect overfitting and model performance. We are also interested in knowing whether and how learning rate and decay interact to affect overfitting and prediction performance, because they are often used together in the SGD optimizer. We also conducted the interactive grid search with batch size and epochs and with L1 and L2, because batch size and epochs are both related to number of data points the model "sees" during the training, and L1 and L2 are both a regularization method that adjusts the loss function. The fixed values of the remaining hyperparameters are the based values as shown in Table 1. The ranges of values used for two of the hyperparameters in the interactive experiments are shown in Table 2.

TABLE 2
HYPERPARAMETERS AND THEIR VALUES IN INTERACTIVE GRID SEARCH

| Paired Hyperparameters | First hyperparameter (number of values) | Second hyperparameter (number of values) |
|---|---|---|
| Learning Rate and Momentum | 0.01- 0.2 with the step size 0.01 (20) | 0.1- 0.9 with the step size 0.005 (9) |
| Learning Rate and Iteration Decay | 0.01- 0.2 with the step size 0.01 (20) | 0.005- 0.1 with the step size 0.005 (20) |
| Batch Size and Epochs | 1-1500 at a step size of 50 (30) | 5-500 at a step size of 25 (20) |
| L1 and L2 | 0 - 0.01 with the step size 0.001 and 0.02 (12) | 0 - 0.01 with the step size 0.001 and 0.02 (12) |

### C. Evaluation Metrics and Dataset

#### 1) Measurement for prediction performance

For a given binary diagnostic test, a receiver operator characteristic (ROC) curve plots the true positive rate against the false positive rate for all possible cutoff values. The Area Under a ROC Curve (AUC) measures the discrimination performance of a model [23]. The higher the AUC, the better the performance of the model at distinguishing between the positive and negative classes. In normal case, the valid AUC should be between 0.5 to 1 which means this model will be able to distinguish different classes. When the AUC is equal to 0.5, the classifier would overall have the poorest performance in terms of distinguishing positive and negative classes [32]. AUC is one of the most important metrics for evaluating the classification model performance, and it has been traditionally used in medical diagnosis since the 1970s [33].

In this research, the deep neural network was trained on the LSM 5-year dataset that was published in previous studies. This dataset contains 4189 patient cases and 31 clinical features that are used as the predictors by the FNN models. The class feature is a binary variable representing whether a patient metastasized within 5 years of the initial treatment. Please refer to our supplementary table for a detailed description of all the variables included in this dataset.

We developed a custom grid search output format that documents 64 output values for each model trained using grid search, including both the results and computer system information, model performance numbers, and total computation time. Each binary diagnostic test outputs a receiver operator characteristic (ROC) curve, a plot of the true positive rate vs. the false positive rate for cutoff values. The area under said ROC curve computes the model's discrimination performance. This area under ROC, also known as AUC, has been extensively used in medical diagnosis since the 1970s and is still one of the crucial methods used to judge classification performance in machine learning and deep learning models [32]-[33]. To compute the AUC of our deep learning model we used the 5-fold cross validation technique to equally split the dataset into five portions for training and testing. This severance was almost fully random, except we had to ensure that each fraction of the dataset was accurately represented so around twenty percent of both positive and negative cases were assigned to each portion. We trained and tested the model five separate times, and each iteration a different portion acted as the validation data set to assess the model which was trained on the other four portions. AUC values for both training and testing were recorded for each of the five train/test cycles alongside the average AUC values over all five cycles. We used the best mean test AUC to choose the final hyperparameter values for this study, and the optimized model would include all these best hyperparameter values. The procedure outlined above was used for all experiments and methods done in this study.

#### 2) Measurement for overfitting

We used percent_AUC_diff to measure overfitting, a measurement which we introduced in our previous study [23]. It represents the percent difference of the average AUC of the 5 training sets and the average AUC of the 5 testing sets during the 5-fold CV process. The average AUC for training, denoted as mean_train_AUC, and the average AUC for testing is denoted as mean_test_AUC, are both part of the standard output values of a grid search. The specific formula for computing percent_AUC_diff is given as percent_AUC_diff = (mean_train_AUC – mean_test_AUC) / mean_test_AUC. The mean_train_AUC is expected to be somewhat better than the mean_test_AUC because models are trained by the training sets of data (so called the cyclic effect). But when the average AUC for the training is significantly higher than the average AUC for testing, or when the percent_AUC_diff is higher than a threshold value such as 5%, we can consider that the model is overfitted.

## IV. RESULTS

Recall that for each of the hyperparameter we conducted our unique type of grid search 30 times, and each time we obtained the 5-fold cross validation results for each of the models trained during the grid search, including the mean_train_AUC, mean_test_AUC, and percent_AUC_diff. The mean_test_AUC is the mean of testing AUCs of the 5 independent testing sets during a 5-fold cross validation and used to access the prediction performance of a model. Mean_train_AUC is the mean of training AUCs of the 5 training sets during a 5-fold cross validation. Percent_AUC_diff is used to quantify overfitting and computed based on both mean_train_AUC and mean_test_AUC. The mean_train_AUCs, mean_test_AUCs, and the percent_AUC_diffs shown in our result Fig.s are the averaged values of the 30 experiments, so they are means of means. Fig. 4 is a panel of 6 Figs that shows our averaged grid search results for the hyperparameters activation function, weight initializer, and number of hidden layer. Fig. 4a) is a bar chart showing the average mean_train_AUCs, and mean_test_AUCs, of the normal sigmoid and hard sigmoid, the two values we tested for the activation function of the output layer. Fig. 4b) shows the matching average percent_AUC_diff for these two values. Hard Sigmoid is piecewise linear approximation of the Logistic Sigmoid activation function which was designed to reduce the calculation time of normal sigmoid activation problem [34]-[35]. Both normal sigmoid and hard sigmoid are suitable for the binary output that we have. Based on Fig. 4a) and 4b), the normal sigmoid performs better than hard sigmoid in terms of both mean_train_AUC and mean_test_AUC, but the average percent_AUC_diff of the normal sigmoid is slightly higher than that of the hard sigmoid. Weight initialization method of a neural network has a vital influence on the convergence speed and performance of the model [36]. A good weight initialization can help alleviate the problem of gradient disappearance and gradient explosion [37]. We tested 5 values: uniform, normal, Glorot_normal, Glorot_uniform, and Lecun_uniform. As shown in Fig. 4c), the normal, Glorot_normal, Glorot_uniform, and Lecun_uniform performed better the two uniform type of weight initializer in terms the mean_train_AUC and mean_test_AUC. In terms of overfitting, as shown in Fig. 4d), the regular uniform and normal weight initializers are doing better than the other three with lower percent_AUC_diff values. We also included the number of hidden layers as one of the hyperparameters, because based on some previous studies, in most cases, the more layers a model has, the more complex the model is, and the more likely model is doing worse in terms of overfitting [38]-[39]. Based on our results as shown in Fig. 4e), the 2-hidden-layers models perform slightly better and the 4-hidden-layers models perform slightly worse than the other models, but overall all types of models perform similarly. It is worth noting that based on our results in Fig. 4f), it is not necessarily true that a model with more layers is doing worse in terms of overfitting.

Results concerning learning rate, momentum, decay, and dropout appear in Fig. 5. Learning rate is used to update weights in the gradient descent procedure during training. According to a study described in [20], the lower the learning rate, the slower the gradient decreases, and the more easily for the model to overfit. Besides, learning rate is a critical hyperparameter for striking a balance between elongated convergence time and not converging at all due to gradient explosion [12]. Our Fig. 5a) shows the average changes of model performance in terms of mean_test_AUC and mean_train_AUC when learning rate gradually increases. Fig. 5b) shows how overfitting changes with learning rate. Based on Fig. 5, we tend to obtain the best performing model when learning rate is between 0.05 and 0.1, because when learning rate is at this range, the mean_test_AUC stays high while overfitting drops down quickly. According to Fig. 5b), when learn rate is low, the model tends to overfit, and this result is consistent with what was reported in [20]. But we do notice that overfitting is in fact steadily getting higher before learning late reaches 0.05, and then it starts to decrease once learn rate surpasses 0.05. Momentum is used to add a fraction of preceding weight update to the current weight update to help dampen gradient oscillation when gradient keeps changing direction. But when gradient persistently points to a certain direction for an extended period with a lot of momentum, training can be trapped to a local minimum, which can lead to overfitting. This explains our results in Fig. 5d), which demonstrates that model overfitting continuously gets worse when the momentum increases. In addition, based on Fig. 5c) and d) when the momentum exceeds 0.5, we tend to obtain a good model which has high mean_test_AUC without being too much overtrained. Decay modifies a model's loss function in a way that allows it to phase out internal weights that have not been updated recently. This helps reducing the complexity of a model while not affecting performance [41]. Dropout can also help in reducing the complexity of a model and therefore reducing overfitting, and it does this by dropping out nodes randomly based on a predetermined probability [42]. Both the performance curves in Fig. 5e) and the overfitting curve in Fig. 5f) for decay show a negative correlation. Overall, when decay increases in value, overfitting will decrease but mostly at the cost of degrading prediction performance. However, we notice that when decay falls in a small range of values, approximately from 0.005 to 0.01, overfitting drops more effectively without hurting the performance as much. This may indicate that we can identify a good model more efficiently when focusing on this small range of values of decay during grid search. The situation for dropout is a bit more complicated. As shown in Fig. 5h), when dropout rate is below 0.5, it tends to be negatively correlated with overfitting, but when it is greater than 0.5, it becomes mostly positively correlated with overfitting. Fig. 5g) shows that the prediction performance is negatively correlated with dropout, and this correlation becomes especially strong when dropout rate reaches 0.8. Combining both Fig.s, we can see that we more likely obtain a good model when dropout rate is less than 0.5.

Fig. 6 contains results for batch size, epochs, L1, and L2. Dividing the data into smaller batches and passing one batch to the model at a time is a strategy of improving training efficiency [45]. Batch size is a hyperparameter that determines how many data points the model sees at any given time. The values of batch size we used ranges from 1 to the maximum number of data points contained in the dataset (4186) with a step size of 5. Fig. 6a) and 6b) shows that both prediction performance and overfitting are negatively correlated with batch size, but overfitting is not significant (less than the 5% threshold) even in the worst case. Based on this result, we tend to obtain the best performing model when batch size is very small. A possible explanation is that a large batch size feeder tends to converge to sharp minimizers of the training and testing functions and that sharp minima lead to poorer generalization. Epochs is the number of times when the entire data set is seen by the model. When epoch is one, the model is trained by each of the samples in the entire training data exactly once [43]. When number of epochs is too low, the model is not trained sufficiently, which causes underfitting. But as the number of epochs increases, the number of weight updates increases, and the chance of overfitting also increases [44]. According to Fig. 6c) and d), both prediction performance and overfitting are positively correlated with epochs, while the performance improvement apparently slows down when the number of epochs exceeds 250. L1 and L2 regularization methods are commonly used in machine learning to control model complexity and reduce overfitting [46]-[47]. Based on Fig. 6e) and f), we obtain the best performing model before L1 reaches about 0.03. After that, as L1 increases, the prediction performance degrades quickly and overfitting tends to become slightly worse instead of better. Fig. 6g) shows that overfitting is negatively correlated with L2 throughout, and Fig. 6h) demonstrates that prediction performance is positively correlated with L2 when L2 is below 0.1, but it become negatively correlated with L2 when L2 exceeds 0.1. Overall, we tend to obtain the best performing model before L2 reaches 0.1, when prediction performance has not yet been degraded but overfitting is already in the tolerable range.

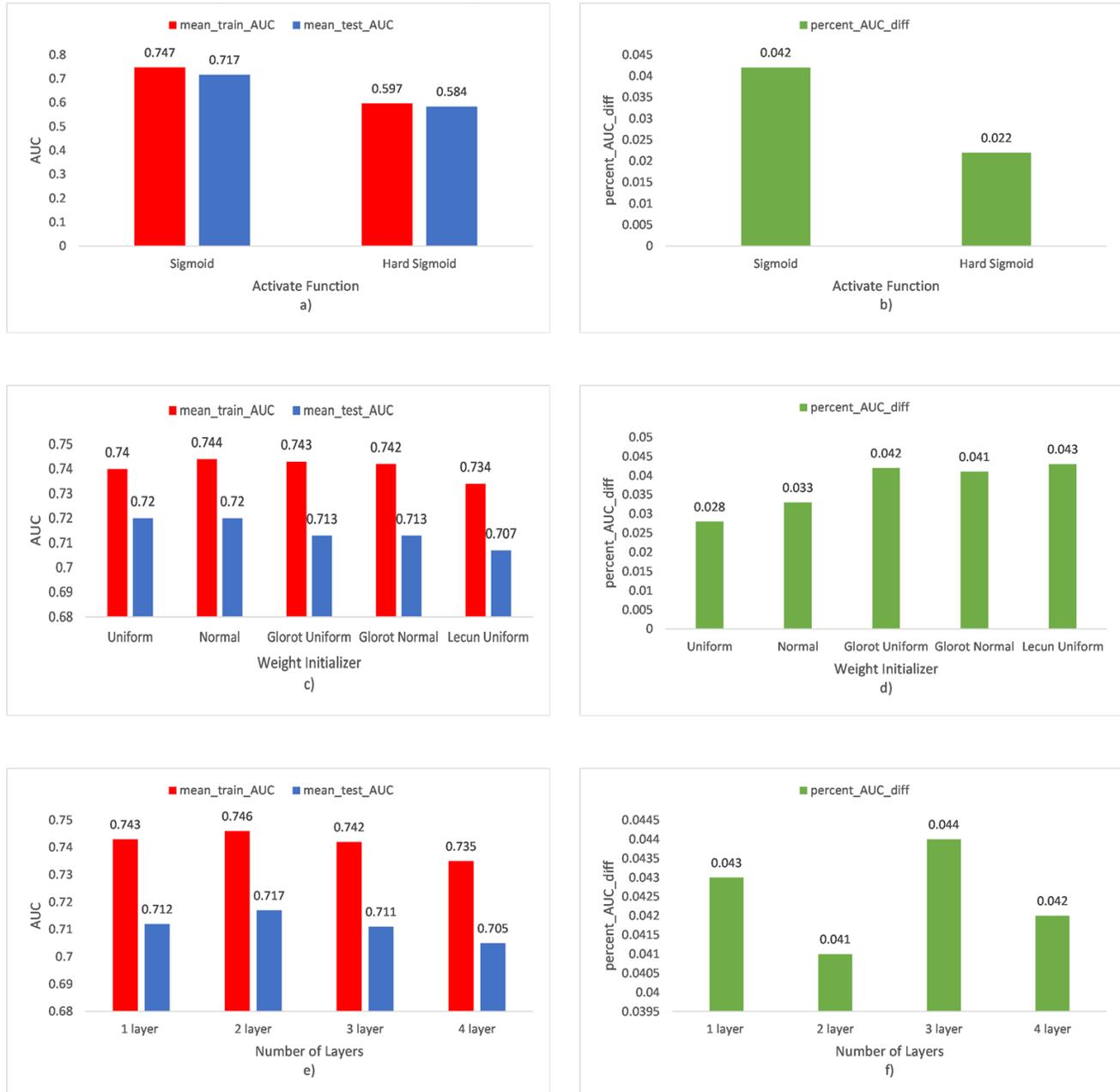

Fig. 4. Average mean_train_AUC, mean_test_AUC, and percent_AUC_diff for activation function, weight initializer, and number of layers.

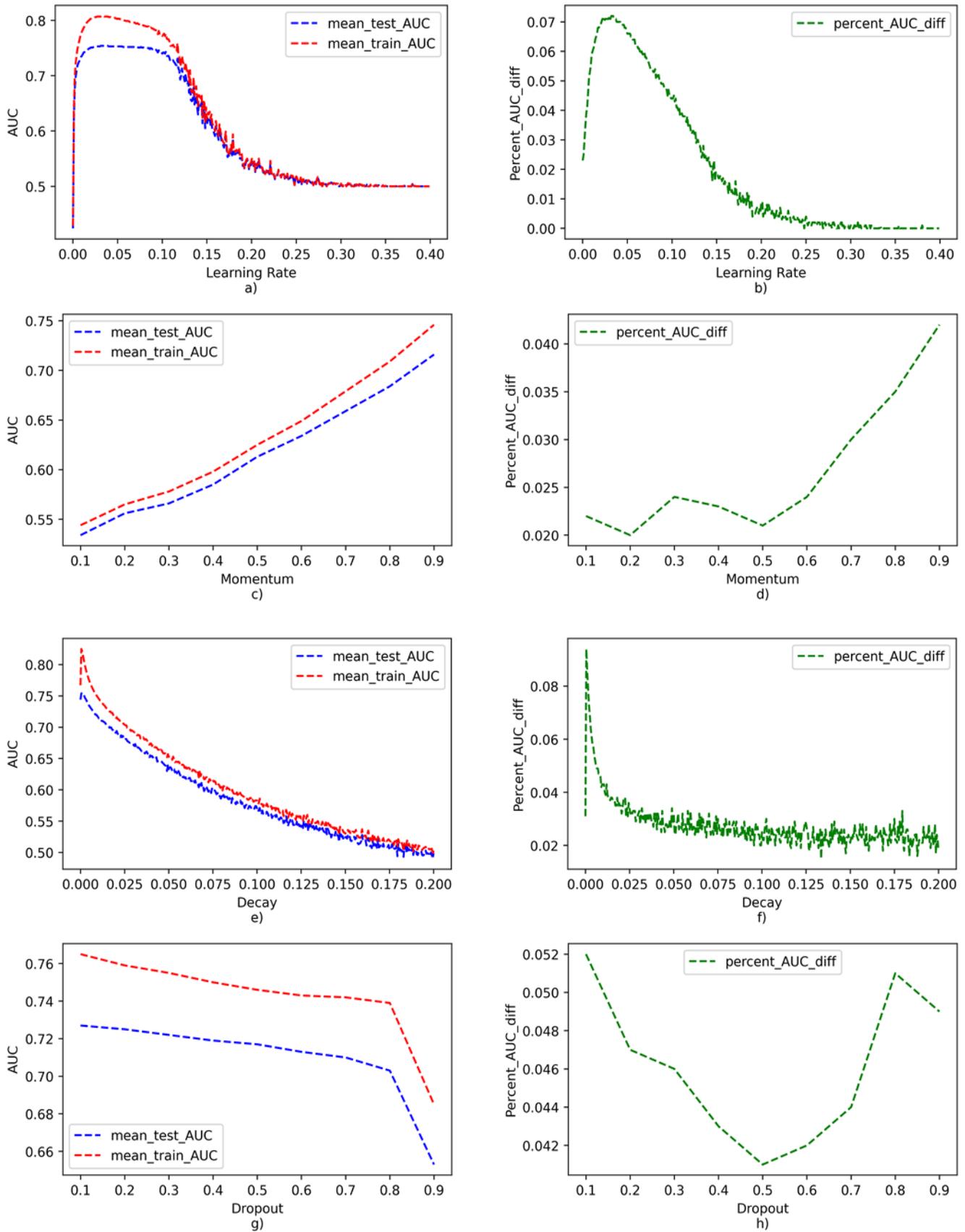

Fig. 5. Average mean_train_AUC, mean_test_AUC, and percent_AUC_diff for learning rate, momentum, decay, and dropout.

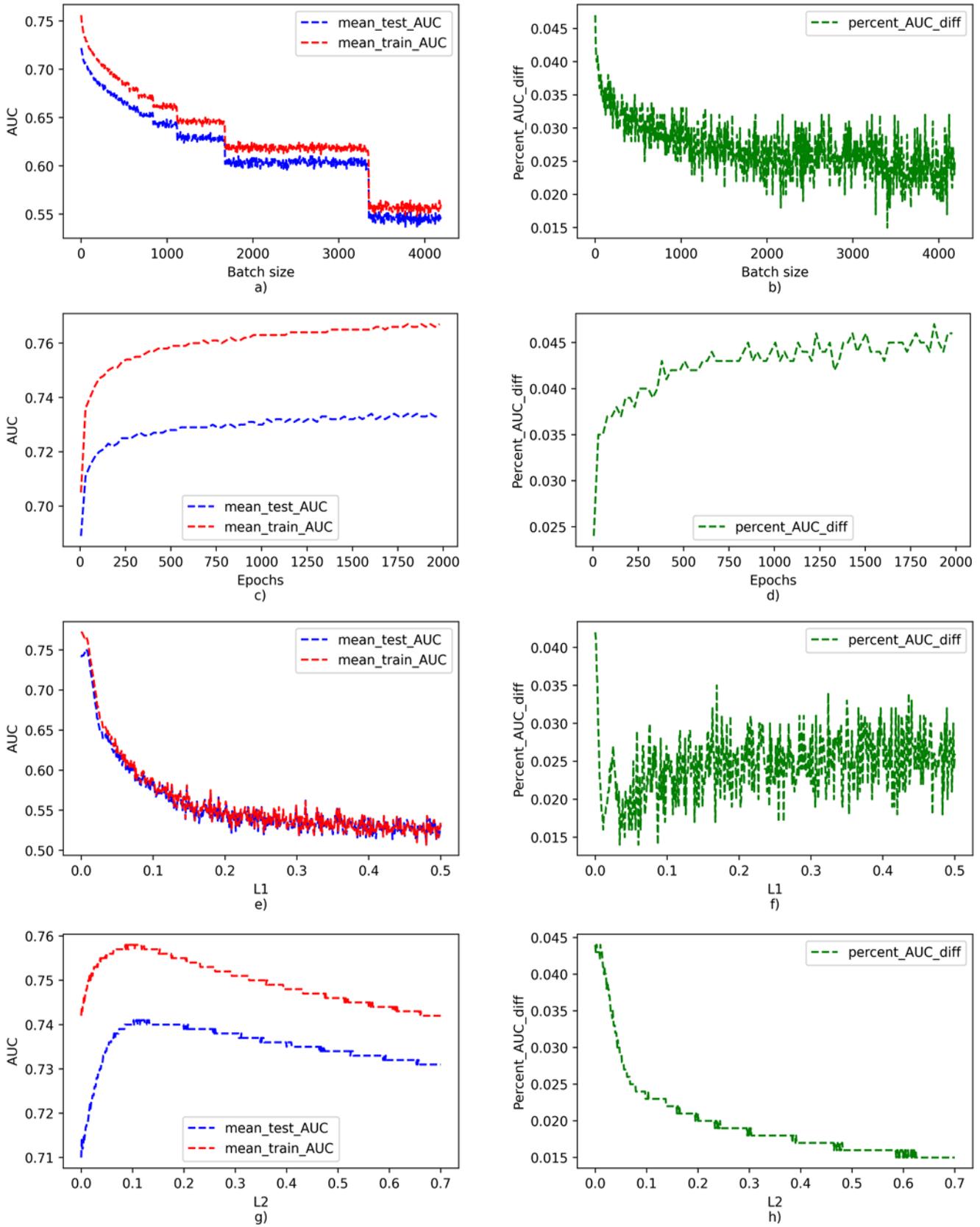

Fig. 6. Average mean_train_AUC, mean_test_AUC, and percent_AUC_diff for batch size, epochs, L1, and L2.

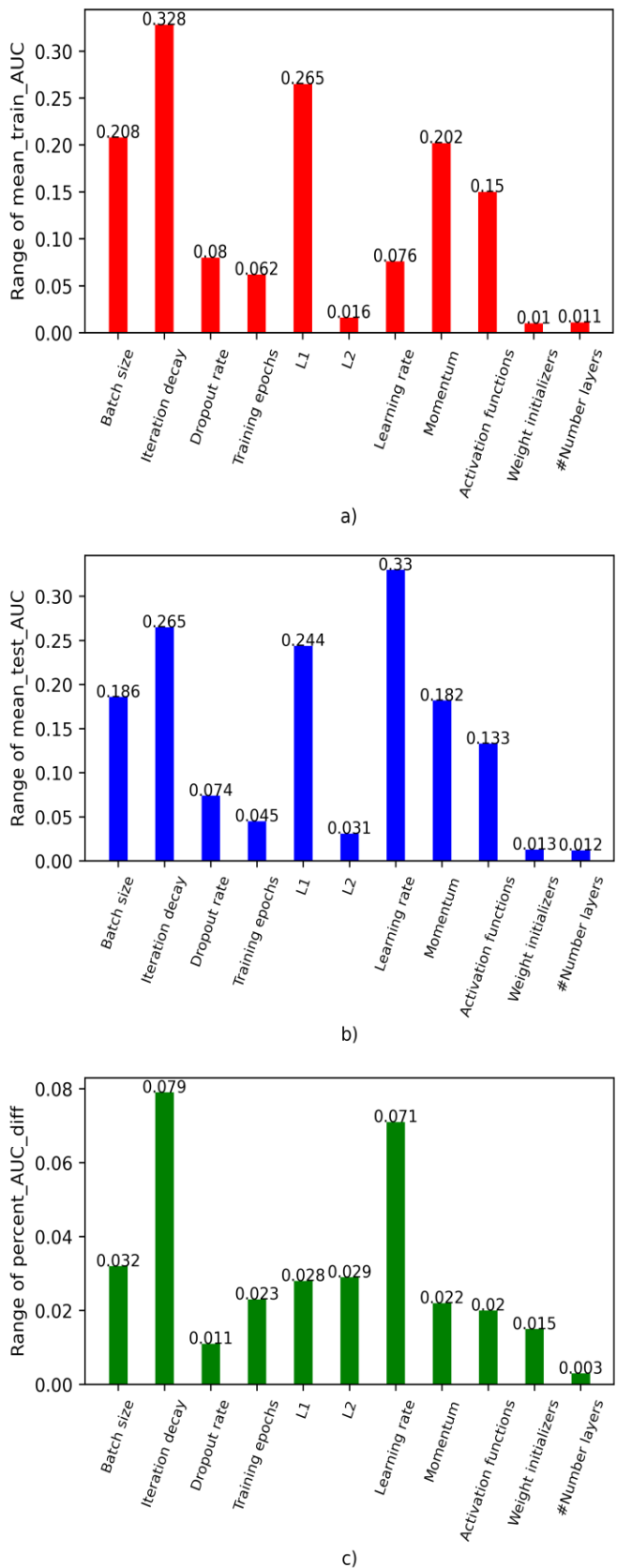

Fig. 7. Ranges of the mean_train_AUCs, mean_test_AUCs, and percent_AUC_diffs of the Hyperparameters.

For every grid search we ran for each of the hyperparameters, we obtained the maximum and the minimum values of the mean_train_AUC, mean_test_AUC, and percent_AUC_diff from all the models trained. We then computed the averaged maximums and minimums of the 30 grid searches for each of the hyperparameters. The ranges of mean_train_AUCs, mean_test_AUC, and percent_AUC_diffs for each of the hyperparameters are the differences between the averaged maximums and minimums, which are shown respectively in Fig. 7a), 7b), and 7c).

According to Fig7. a), the top three hyperparameters in terms of the range of mean_train_AUC are decay (1st, 0.328), L1 (2nd, 0.265), and batch size (3rd, 0.208). It is not hard to observe thar there is a strong correlation between Fig. 7a) and Fig. 7b) except hyperparameter learning rate which has huge range on mean_test_AUC but relatively small range on mean_train_AUC. The bottom three hyperparameters are same for both mean_test_AUC and mean_train_AUC including weight initializer, number of hidden layer and L2.

Based on Fig. 7c), the top three hyperparameters in terms of the range of percent_AUC_diff are decay (1st, 0.079), learning rate (2nd, 0.071), and batch size (3rd, 0.032). The bottom three are number of hidden layers (11th, 0.003), Dropout Rate (10th, 0.011) and Weight Initializer (9th, 0.015). Based on Fig. 7b), the top three hyperparameters in terms of the range of mean_test_AUC are Learning rate (1st, 0.33,)), decay (2nd, 0.265) and L1 (3rd, 0.244), and the bottom three are number of hidden layers (11th, 0.012), weight initializer (10th, 0.013), and L2 (9th, 0.031).

Based on Fig. 7, changing values of learning rate, decay, and batch size has a more significant impact on both overfitting and prediction performance than doing so with most of the other hyperparameters, including the ones that were designed to for the purpose of minimizing overfitting such as L1, L2, and dropout. Overfitting is reported to be associated with a large number of hidden layers [21], but based our results, overfitting is the least sensitive to number of hidden layers.

The results of our paired hyperparameter experiments are shown in Fig. 8. It is believed that a setup where learning rate and momentum have a negative correlation is the best [40]. A performance drop is indeed "seen" in Fig. 8 a) when a lot of momentum combines with large learning rates. Based on Fig. 8a) and b), we obtained the best mean_test_AUCs but high overfitting when small learning rates combines with a lot of momentum. Fig. 8c) and d) show that a combination of large learning rates and large values of decay may help improve results, while a large learning learn can result in poor prediction performance when decay is low. Fig. 8e) and f) show that although predict perform overall negatively corelates with batch size, but it can be "lifted" up when a large batch size meets with a large number of epochs. Interestingly, Fig. 8g) and h) seem to indicate that L2 alone does not have a very high impact on either the prediction performance or overfitting, neither does it seem to interact much with L1.

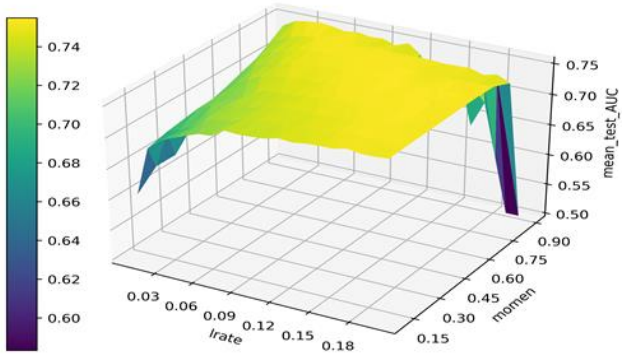

a) Learning rate, momentum, and mean_test_AUC

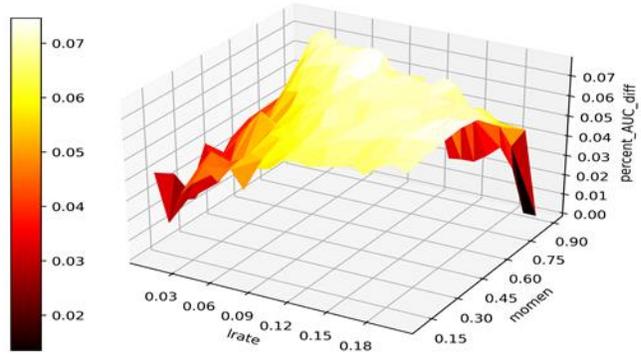

b) Learning rate, momentum, and percent_AUC_diff

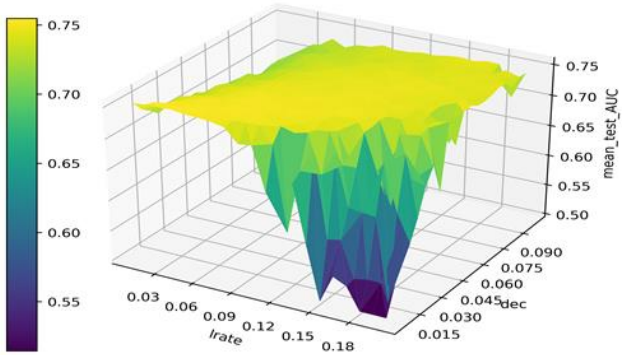

c) Learning rate, decay, and mean_test_AUC

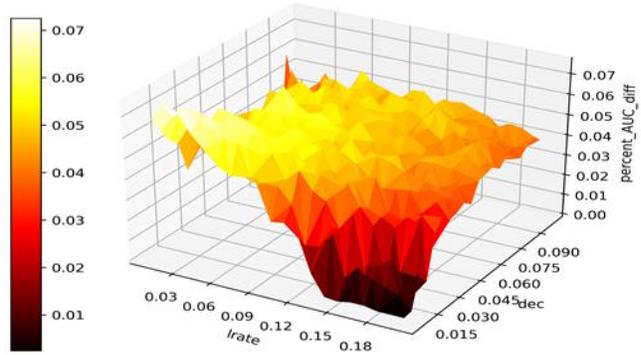

d) Learning rate, decay, and percent_AUC_diff

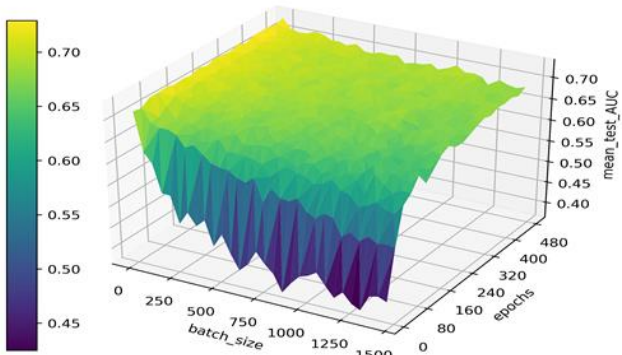

e) Batch size, epochs, and mean_test_AUC

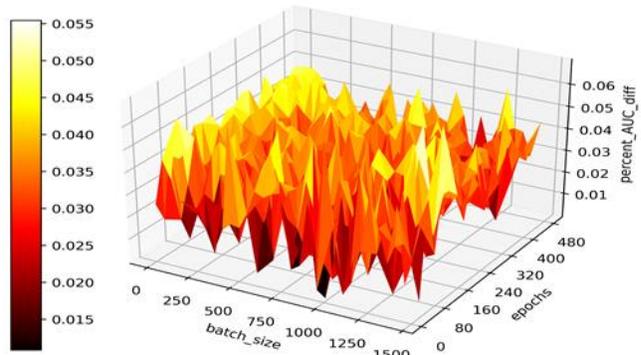

f) Batch size, epochs, and percent_AUC_diff

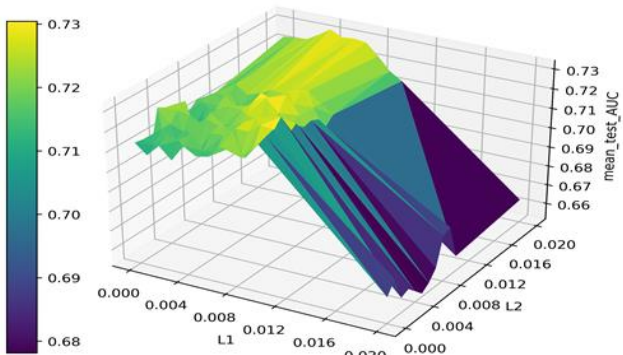

g) L1, L2, and mean_test_AUC

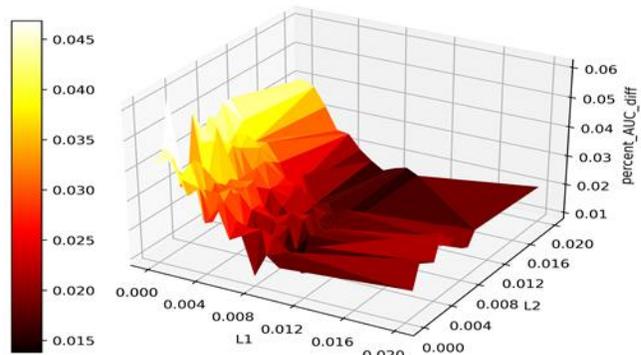

h) L1, L2 and percent_AUC_diff

Fig. 8. Results of the interactive experiments with three different hyperparameter pairs.

## V. Conclusion

Our results not only substantiated some of the existing knowledge in the field of machine learning but also presented interesting new findings. More specifically, our results are consistent with some of the existing research finding or knowledge such as activation function is associated with overfitting [20], the predict performance tends to drop when a lot of momentum works together with a large learning rate [40], and smaller batch size is often associated with better prediction performance. On the other hand, we expected to see that the number of layers is closely associated with overfitting based on literature [21], but this does not show clearly in our results. In addition, our results show that most of the single hyperparameters are either negatively or positively corrected with model prediction performance and overfitting. In particular, we found that overfitting overall tends to negatively correlate with learning rate, decay, batch sides, and L2, but tends to positively correlate with momentum, epochs, and L1. As discussed in the results section, we also noticed that prediction models are prone to perform better within a certain range of hyperparameter values for some of the hyperparameters. For example, we are more likely to see better results when momentum exceeds 0.5, when drop out is below 0.5, or before L1 reaches 0.02. These types of findings are useful for selecting the range of hyperparameter values required by grid search. According to our results, learning rate, decay, and batch size may have a more significant impact on both overfitting and prediction performance than most of the other hyperparameters, including the ones that were designed for the purpose of minimizing overfitting such as L1, L2, and dropout.

## Appendix

| Variables included | Description | Values |
|---|---|---|
| race | race of patient | white, black, Asian, American Indian or Alaskan native, native Hawaiian or other pacific islander |
| ethnicity | ethnicity of patient | not Hispanic, Hispanic |
| smoking | smoking history of patient | ex smoker, non smoker, cigarettes, chewing tobacco, cigar |
| alcohol usage | alcohol usage of patient | moderate, no use, use but nos, former user, heavy user |
| family history | family history of cancer | cancer, no cancer, breast cancer, other cancer, cancer but nos |
| age_at_diagnosis | age at diagnosis of the disease | 0-49, 50-69, >69 |
| menopausal_status | inferred menopausal status | pre, post |
| side | side of tumor | left, right |
| TNEG | patient ER, PR, and HER2 negative | yes, no |
| ER | estrogen receptor expression | neg, pos, low pos |
| ER_percent | percent of cell stain pos for er receptors | 90-100, 0-20, 20-90 |
| PR | progesterone receptor expression | neg, pos, low pos |
| PR_percent | percent of cell stain pos for pr receptors | 90-100, 0-20, 20-90 |
| P53 | whether P53 is mutated | neg, pos, low pos |
| HER2 | HER2 expression | neg, pos |
| t_tnm_stage | prime tumor stage in tnm system | 0, 1,2,3,4, IS, 1mic, X |
| n_tnm_stage | # nearby cancerous lymph nodes | 0,1,2,3,4,X |
| stage | composite of size and # positive nodes | 0,1,2,3 |
| lymph_nodes_removed | number of lymph nodes removed | 0-11, 12-22, > 22 |
| lymph_nodes_positive | number of positive lymph nodes | 0, 1-8 >8 |
| lymph_node_status | patient had any positive lymph nodes | neg,pos |
| histology | tumor histology | lobular, duct |
| size | size of tumor in mm | 0-32, 32-70, >70 |
| grade | grade of disease | 1, 2, 3 |
| invasive | whether tumor is invasive | yes,no |
| histology2 | tumor histology subtypes | IDC, DCIS, ILC, NC |
| invasive_tumor_location | where invasive tumor is located | mixed duct and lobular, duct, lobular, none |
| DCIS_level | type of ductal carcinoma in situ | solid, apocrine, cribriform, dcis, comedo, papillary, micropapillary |
| re_excision | removal of an additional margin of tissue | yes, no |
| surgical_margins | whether residual tumor | res. tumor, no res. tumor, no primary site surgery |
| MRIs_60_surgery | | yes, no |

## Acknowledgment

PCP was involved in the work as a trainee in XJ's Artificial Intelligence research lab via UPMC Hillman Cancer Center Academy.

## References

[1] H. Sung *et al.*, "Global Cancer Statistics 2020: GLOBOCAN Estimates of Incidence and Mortality Worldwide for 36 Cancers in 185 Countries," *CA. Cancer J. Clin.*, vol. 71, no. 3, pp. 209–249, May 2021, doi: 10.3322/CAAC.21660.

[2] L. Rahib, M. R. Wehner, L. M. Matrisian, and K. T. Nead, "Estimated Projection of US Cancer Incidence and Death to 2040," *JAMA Netw. Open*, vol. 4, no. 4, Apr. 2021, doi: 10.1001/JAMANETWORKOPEN.2021.4708.

[3] "Cancer Facts & Figures 2021 | American Cancer Society." https://www.cancer.org/research/cancer-facts-statistics/all-cancer-facts-figures/cancer-facts-figures-2021.html (accessed Dec. 02, 2021).

[4] C. E. DeSantis *et al.*, "Breast cancer statistics, 2019," *CA. Cancer J. Clin.*, vol. 69, no. 6, pp. 438–451, Nov. 2019, doi: 10.3322/CAAC.21583.

[5] A. M. Afifi, A. M. Saad, M. J. Al-Husseini, A. O. Elmehrath, D. W. Northfelt, and M. B. Sonbol, "Causes of death after breast cancer diagnosis: A US population-based analysis," *Cancer*, vol. 126, no. 7, pp. 1559–1567, Apr. 2020, doi:

10.1002/CNCR.32648.
[6]  R. L. Siegel, K. D. Miller, and A. Jemal, "Cancer statistics, 2020," *CA. Cancer J. Clin.*, vol. 70, no. 1, pp. 7–30, Jan. 2020, doi: 10.3322/CAAC.21590.
[7]  G. P. Gupta and J. Massagué, "Cancer Metastasis: Building a Framework," *Cell*, vol. 127, no. 4, pp. 679–695, Nov. 2006, doi: 10.1016/J.CELL.2006.11.001.
[8]  I. Saritas, "Prediction of Breast Cancer Using Artificial Neural Networks," *J. Med. Syst.*, vol. 36, no. 5, pp. 2901–2907, Oct. 2012, doi: 10.1007/S10916-011-9768-0.
[9]  L. Ran, Y. Zhang, Q. Zhang, and T. Yang, "Convolutional Neural Network-Based Robot Navigation Using Uncalibrated Spherical Images," *Sensors 2017, Vol. 17, Page 1341*, vol. 17, no. 6, p. 1341, Jun. 2017, doi: 10.3390/S17061341.
[10] B. Weigelt, F. L. Baehner, and J. S. Reis-Filho, "The contribution of gene expression profiling to breast cancer classification, prognostication and prediction: a retrospective of the last decade," *J. Pathol.*, vol. 220, no. 2, pp. 263–280, Jan. 2010, doi: 10.1002/PATH.2648.
[11] S. Belciug and F. Gorunescu, "A hybrid neural network/genetic algorithm applied to breast cancer detection and recurrence," *Expert Syst.*, vol. 30, no. 3, pp. 243–254, Jul. 2013, doi: 10.1111/J.1468-0394.2012.00635.X.
[12] S. Lawrence and C. L. Giles, "Overfitting and neural networks: Conjugate gradient and backpropagation," *Proc. Int. Jt. Conf. Neural Networks*, vol. 1, pp. 114–119, 2000, doi: 10.1109/IJCNN.2000.857823.
[13] Z. Li, K. Kamnitsas, and B. Glocker, "Overfitting of Neural Nets Under Class Imbalance: Analysis and Improvements for Segmentation," *Lect. Notes Comput. Sci. (including Subser. Lect. Notes Artif. Intell. Lect. Notes Bioinformatics)*, vol. 11766 LNCS, pp. 402–410, Oct. 2019, doi: 10.1007/978-3-030-32248-9_45.
[14] IBM Cloud Education, "What is Underfitting?," *IBM*, Mar. 21, 2021. https://www.ibm.com/cloud/learn/underfitting#toc-ibm-and-un-6BYka0Vn (accessed Jun. 30, 2022).
[15] Wi. Koehrsen, "Overfitting vs. Underfitting: A Complete Example," *Towards Data Science*, Jan. 28, 2018. https://towardsdatascience.com/overfitting-vs-underfitting-a-complete-example-d05dd7e19765 (accessed Jun. 30, 2022).
[16] M. Decuyper *et al.*, "An Overview of Overfitting and its Solutions," *J. Phys. Conf. Ser.*, vol. 1168, no. 2, p. 022022, Feb. 2019, doi: 10.1088/1742-6596/1168/2/022022.
[17] R. B. Arif, M. A. B. Siddique, M. M. R. Khan, and M. R. Oishe, "Study and observation of the variations of accuracies for handwritten digits recognition with various hidden layers and epochs using convolutional neural network," *4th Int. Conf. Electr. Eng. Inf. Commun. Technol. iCEEiCT 2018*, pp. 112–117, Sep. 2018, doi: 10.1109/CEEICT.2018.8628078.
[18] "On Dropout, Overfitting, and Interaction Effects in Deep Neural Networks | OpenReview." https://openreview.net/forum?id=68747kJ0qKt (accessed Jun. 30, 2022).
[19] O. Demir-Kavuk, M. Kamada, T. Akutsu, and E. W. Knapp, "Prediction using step-wise L1, L2 regularization and feature selection for small data sets with large number of features," *BMC Bioinformatics*, vol. 12, no. 1, pp. 1–10, Oct. 2011, doi: 10.1186/1471-2105-12-412/TABLES/2.
[20] H. Li, J. Li, X. Guan, B. Liang, Y. Lai, and X. Luo, "Research on Overfitting of Deep Learning," *Proc. - 2019 15th Int. Conf. Comput. Intell. Secur. CIS 2019*, pp. 78–81, Dec. 2019, doi: 10.1109/CIS.2019.00025.
[21] H. Il Suk, "An Introduction to Neural Networks and Deep Learning," *Deep Learn. Med. Image Anal.*, pp. 3–24, Jan. 2017, doi: 10.1016/B978-0-12-810408-8.00002-X.
[22] S. Li *et al.*, "Deep Learning for Hyperspectral Image Classification: An Overview," Accessed: Jun. 29, 2022. [Online]. Available: http://www.webofknowledge.com/WOS.
[23] X. Jiang and C. Xu, "Improving Clinical Prediction of Later Occurrence of Breast Cancer Metastasis Using Deep Learning and Machine Learning with Grid Search," Jun. 2022, doi: 10.20944/PREPRINTS202206.0394.V1.
[24] "Dropout: A Simple Way to Prevent Neural Networks from Overfitting." https://jmlr.org/papers/v15/srivastava14a.html (accessed Jul. 08, 2022).
[25] "SGD: General Analysis and Improved Rates." http://proceedings.mlr.press/v97/qian19b (accessed Jul. 08, 2022).
[26] S. Tschiatschek, K. Paul, and F. Pernkopf, "Integer Bayesian network classifiers," *Lect. Notes Comput. Sci. (including Subser. Lect. Notes Artif. Intell. Lect. Notes Bioinformatics)*, vol. 8726 LNAI, no. PART 3, pp. 209–224, 2014, doi: 10.1007/978-3-662-44845-8_14.
[27] "Control Batch Size and Learning Rate to Generalize Well: Theoretical and Empirical Evidence." https://proceedings.neurips.cc/paper/2019/hash/dc6a70712a252123c40d2adba6a11d84-Abstract.html (accessed Jul. 08, 2022).
[28] D. Wang, A. Khosla, R. Gargeya, H. Irshad, and A. H. Beck, "Deep Learning for Identifying Metastatic Breast Cancer," Jun. 2016, Accessed: Aug. 05, 2021. [Online]. Available: https://arxiv.org/abs/1606.05718v1.
[29] A. Nih, "The Precision Medicine Initiative Cohort Program – Building a Research Foundation for 21st Century Medicine," 2015.
[30] X. Jiang, A. Wells, A. Brufsky, and R. Neapolitan, "A clinical decision support system learned from data to personalize treatment recommendations towards preventing breast cancer metastasis," *PLoS One*, vol. 14, no. 3, p. e0213292, Mar. 2019, doi: 10.1371/JOURNAL.PONE.0213292.
[31] X. Jiang, A. Wells, A. Brufsky, D. Shetty, K. Shajihan, and R. E. Neapolitan, "Leveraging Bayesian networks and information theory to learn risk factors for breast cancer metastasis," *BMC Bioinformatics*, vol. 21, no. 1, pp. 1–17, Jul. 2020, doi: 10.1186/S12859-020-03638-8/FIGURES/5.
[32] J. Huang and C. X. Ling, "Using AUC and accuracy in evaluating learning algorithms," *IEEE Trans. Knowl. Data Eng.*, vol. 17, no. 3, pp. 299–310, Mar. 2005, doi: 10.1109/TKDE.2005.50.
[33] J. Brownlee, "How to Grid Search Hyperparameters for Deep Learning Models in Python With Keras," Accessed: Jun. 29, 2022. [Online]. Available: https://machinelearningmastery.com/grid-search-hyperparameters-deep-learning-models-python-keras/.
[34] P. Ramachandran, B. Zoph, and Q. V Le Google Brain, "Searching for Activation Functions," *6th Int. Conf. Learn. Represent. ICLR 2018 - Work. Track Proc.*, Oct. 2017, Accessed: Dec. 02, 2021. [Online]. Available: https://arxiv.org/abs/1710.05941v2.
[35] C. Gulcehre, M. Moczulski, M. D. Com, Y. Bengio, and B. U. Ca, "Noisy Activation Functions Misha Denil † †," 2016.
[36] S. K. Kumar, "On weight initialization in deep neural networks," Apr. 2017, Accessed: Dec. 02, 2021. [Online]. Available: https://arxiv.org/abs/1704.08863v2.
[37] H. Li, M. Krček, and G. Perin, "A comparison of weight initializers in deep learning-based side-channel analysis," *Lect. Notes Comput. Sci. (including Subser. Lect. Notes Artif. Intell. Lect. Notes Bioinformatics)*, vol. 12418 LNCS, pp. 126–143, 2020, doi: 10.1007/978-3-030-61638-0_8/COVER/.

[38] A. Darmawahyuni *et al.*, "Deep Learning with a Recurrent Network Structure in the Sequence Modeling of Imbalanced Data for ECG-Rhythm Classifier," *Algorithms 2019, Vol. 12, Page 118*, vol. 12, no. 6, p. 118, Jun. 2019, doi: 10.3390/A12060118.

[39] H. Li, J. Li, X. Guan, B. Liang, Y. Lai, and X. Luo, "Research on Overfitting of Deep Learning," in *Proceedings - 2019 15th International Conference on Computational Intelligence and Security, CIS 2019*, Dec. 2019, pp. 78–81, doi: 10.1109/CIS.2019.00025.

[40] N. Schraudolph and F. Cummins, "Momentum and Learning Rate Adaptation," *Introduction to Neural Networks*, 2006. https://cnl.salk.edu/~schraudo/teach/NNcourse/momrate.html (accessed Jun. 30, 2022).

[41] D. Vasani, "This thing called Weight Decay," *Towards Data Science*, Apr. 29, 2019. https://towardsdatascience.com/this-thing-called-weight-decay-a7cd4bcfccab (accessed Jul. 01, 2022).

[42] N. Srivastava, "Improving Neural Networks with Dropout," 2013.

[43] W. Zaremba, I. Sutskever, O. Vinyals, and G. Brain, "Recurrent Neural Network Regularization," Sep. 2014, doi: 10.48550/arxiv.1409.2329.

[44] J. Brownlee, "What is the Difference Between a Batch and an Epoch in a Neural Network?," 2018, Accessed: Jun. 30, 2022. [Online]. Available: https://machinelearningmastery.com/difference-between-a-batch-and-an-epoch/.

[45] D. Mandy, "Batch Size In A Neural Network Explained," *Deeplizard*, Nov. 22, 2017. https://deeplizard.com/learn/video/U4WB9p6ODjM (accessed Jul. 01, 2022).

[46] A. Y. Ng, "L1 and L2 regularisation comparisation," *Proc. 21 st Int. Conf. Mach. Learn.*, 2004.

[47] S. Bekta and Y. Iman, "The comparison of L 1 and L 2-norm minimization methods," Int. J. Phys. Sci., vol. 5, no. 11, pp. 1721–1727, 2010, Accessed: Jun. 30, 2022. [Online]. Available: http://www.academicjournals.org/IJPS